\documentclass[runningheads]{llncs}
\usepackage{graphicx}

\usepackage{comment}
\usepackage{amsmath,amssymb} %
\usepackage{amsfonts}
\usepackage{multirow}
\usepackage{multicol}
\usepackage{overpic}
\usepackage{bbm}

\usepackage[font=footnotesize,labelfont=bf]{caption}
\usepackage[scriptsize]{subfigure}

\usepackage[table]{xcolor}

\newcommand{\ie}{\textit{i.e.}}
\newcommand{\eg}{\textit{e.g.}}
\newcommand{\etc}{\textit{etc}}
\DeclareMathOperator*{\argmax}{arg\,max}
\DeclareMathOperator*{\argmin}{arg\,min}

\usepackage[accsupp]{axessibility}  %

\begin{document}
\pagestyle{headings}
\mainmatter
\def\ECCVSubNumber{3254}  %

\title{Leveraging Action Affinity and Continuity for Semi-supervised Temporal Action Segmentation} %

\titlerunning{Semi-supervised Temporal Action Segmentation}

\author{Guodong Ding\orcidID{0000-0001-6080-5220} \and
Angela Yao\orcidID{0000-0001-7418-6141} }
\authorrunning{G. Ding and A. Yao}
\institute{National University of Singapore\\
\email{\{dinggd, ayao\}@comp.nus.edu.sg}}
\maketitle

\begin{abstract}
We present a semi-supervised learning approach to the temporal action segmentation task. The goal of the task is to temporally detect and segment actions in long, untrimmed procedural videos, where only a small set of videos are densely labelled, and a large collection of videos are unlabelled. To this end, we propose two novel loss functions for the unlabelled data: an action affinity loss and an action continuity loss. The action affinity loss guides the unlabelled samples learning by imposing the action priors induced from the labelled set. Action continuity loss enforces the temporal continuity of actions, which also provides frame-wise classification supervision. In addition, we propose an Adaptive Boundary Smoothing (ABS) approach to build coarser action boundaries for more robust and reliable learning. The proposed loss functions and ABS were evaluated on three benchmarks. Results show that they significantly improved action segmentation performance with a low amount (5\% and 10\%) of labelled data and achieved comparable results to full supervision with 50\% labelled data. Furthermore, ABS succeeded in boosting performance when integrated into fully-supervised learning.
\end{abstract}

\section{Introduction}

Temporal action segmentation aims to segment long, untrimmed procedural video sequences into multiple actions and assign semantic labels for each frame. This task requires the arduous collection of frame-wise labelling for minutes-long videos. Previous works have reduced the annotation effort via weaker supervision in the form of transcripts~\cite{kuehne2017weakly}, action sets~\cite{fayyaz2020sct}, and timestamp labels~\cite{li2021temporal}. With each method, annotators are still required to watch or scrub through each video in the training set to provide labels for \textit{every} training video. 
Different from the previous methods,  we work under a semi-supervised setting where frame-wise annotations are provided for only a small portion (5\% and 10\%) of the videos in the training set, while the remaining videos are unlabelled. This setting greatly reduces the annotation efforts.  

Semi-supervised learning has been studied extensively in image-based vision tasks such as image classification~\cite{rizve2021defense}, object detection~\cite{wang2021data}, semantic segmentation~\cite{he2021re}, \etc. Two popular semi-supervised learning techniques are consistency regularization~\cite{samuli2017temporal,tarvainen2017mean} and pseudo-labelling~\cite{lee2013pseudo}. Consistency regularization assumes that realistic augmentations on the input data will not change the output distribution. Psuedo-labelling generates labels for unlabelled data before training. These techniques are not suitable for video-based tasks for several reasons. Firstly, it is non-trivial to perform realistic data augmentation operations required by consistency regularization methods as the prevailing practice in temporal action segmentation is to use pre-computed feature vectors as input. Furthermore, directly extending na\"ive pseudo-labelling on videos may result in confirmation bias~\cite{arazo2020pseudo}, \ie, overfitting to incorrect pseudo-labels. 

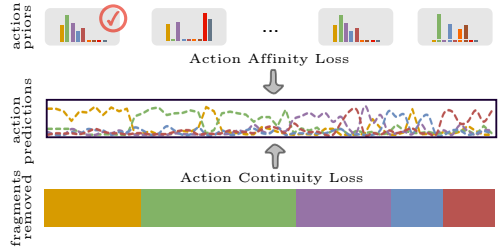
\begin{figure}[tb]
    \centering
    \begin{overpic}[width=0.5\textwidth]{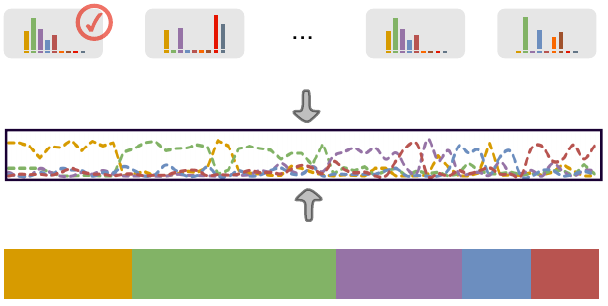}

    \put(32,37){\tiny Action Affinity Loss}
    \put(30,11){\tiny Action Continuity Loss}
    
    \put(-6.8,44){\rotatebox[origin=c]{90}{\tiny action}}
    \put(-4.4,44){\rotatebox[origin=c]{90}{\tiny priors}}
    
    \put(-6.8,23.5){\rotatebox[origin=c]{90}{\tiny action}}
    \put(-4.4,23.5){\rotatebox[origin=c]{90}{\tiny predictions}}

    \put(-6.8,4.5){\rotatebox[origin=c]{90}{\tiny fragments}}
    \put(-4.4,4.5){\rotatebox[origin=c]{90}{\tiny removed}}
    \end{overpic}
    \caption{Overview of our two complementary loss functions. The Action Affinity loss imposes the best matched (denoted by a check mark) prior of action compositions and distributions from the labelled data. The Action Continuity Loss removes the fragments of action labels.}
    \label{fig:teaser}
\end{figure}

Constructing conceivable supervision for unlabelled data in temporal action segmentation task raises questions such as \textit{``What action compositions are likely to occur?", ``What is a reasonable temporal proportion for each action to take?"} and \textit{``What kind of constraints should the action labels follow?" }
We propose to tackle these questions by leveraging two unique observations we made on procedural videos: 1) \textbf{Action Affinity}, procedural videos performing a specific activity (\eg,~`making\_coffee') comprise correlated action units (\eg,~`take\_cup', `pour\_coffee', `pour\_milk', and `stir\_coffee') and there exist pairs of videos that have resembling temporal portions for action unit pairs; 
2) \textbf{Action Continuity}, action labels stay locally constant and action labels only transit at true boundaries. The former is an observation on relations between video instances, while the latter is a video-wise trait. In this work, we propose two novel unsupervised loss functions, action affinity loss and action continuity loss, each leveraging one of these observations. An overview of our losses is depicted in Fig.~\ref{fig:teaser}. 

As opposed to previous work~\cite{li2021temporal}, which uses sparse per-segment labelled frames for \textit{every} video, we use datasets that consist of a small set of densely labelled videos and a large set of unlabelled videos. Dense labels do more than just provide frame-level action labels in the video. They can also be used to establish prior information about the action compositions and distributions. We thus define for a labelled video a high-level representation -- its action frequency, \ie, the temporal proportion of actions. For unlabelled video without action information, we adopt a soft version based on the network predictions.
Action frequency naturally indicates the relative action lengths as it is normalised by video lengths. The fact that it does not constrain the action ordering allows for flexibility as some actions do not necessarily follow a rigid sequence. Considering the action frequency of labelled videos as action priors, we exploit the action affinities between labelled and unlabelled video samples in a heuristic way and integrate them in the action affinity loss function for model training.

Fragmentation or over-segmentation is a common problem in action segmentation~\cite{ishikawa2021alleviating}, which is exacerbated in a semi-supervised setting when networks train and overfit on few labelled samples. To mitigate this, we first propose an action sequence extraction scheme that better captures the underlying action ordering. 
These sequentially sub-sampled actions are later compared against the original network predictions with dynamic time warping~\cite{sakoe1978dynamic} to estimate our action continuity loss. We show that the action continuity loss can take the same form as the classification cross-entropy loss with a proper distance function adopted in dynamic time warping. Although our loss function also aims to maintain the temporal continuity of actions, it differs from the commonly used agnostic smoothing loss of~\cite{farha2019ms}. Ours enforces a specific action ordering while providing frame-wise action supervision.

To add robustness to the action boundaries found by dynamic time warping for the unlabelled videos, we propose a soft transitional boundary. Specifically, we smooth the rigid boundaries so that frames around the boundaries have a mixed probability of belonging to both consecutive action classes. This was previously explored in a weakly-supervised setting~\cite{ding2018weakly}, albeit in a highly rigid form.
In our work, we vary the number of boundary frames depending on the action duration and use a sigmoid function for mixing.

In summary, this paper offers four key contributions:
\begin{enumerate}
    \item By investigating the correlation of actions between labelled and unlabelled procedural videos, we propose an action affinity loss to integrate action priors for semi-supervised learning. 
    \item Building on the continuity property of procedural actions, we propose an action continuity loss to enforce action ordering constraints and provide classification supervision for unlabelled data.
    \item For more robust and reliable learning, we propose a general adaptive boundary smoothing (ABS) technique that generates smoothed coarse action probabilities for boundary frames. Our ABS improves segmentation performance in both semi- and fully-supervised settings.
    \item Experimental results show that our proposed approach improves the segmentation performance by a large margin with a small amount of labelled data (5\% and 10\%) and achieve comparable performance to the fully-supervised setting with 50\% of labelled data. 
\end{enumerate}

\section{Related Work}

\subsection{Temporal Action Segmentation}
Temporal action segmentation in videos has been explored with various levels of supervision.
\textbf{Fully-supervised} approaches require the long videos in the training set to be densely annotated. The input and label pairs are fed into a Temporal Convolution Network (TCN) to learn a mapping to frame-wise action labels~\cite{lea2017temporal,lei2018temporal,farha2019ms,singhania2021coarse}. 
\textbf{Weakly-supervised} methods use action lists~\cite{kuehne2017weakly} or action sets~\cite{richard2018action} to learn the alignment between actions and frames. Specifically, D3TW~\cite{chang2019d3tw} proposed a differentiable dynamic time warping loss with continuous relaxation to discriminatively learn the best alignment when multiple action lists were provided. Compared to their work, we do not use any action list supervision, and we utilise DTW as an optimisation tool for making the best assignment that meets the ordering constraints. A more recent weakly-supervised work~\cite{li2021temporal} learns to segment actions via a small percentage of action timestamps. Although weak supervision reduces the effort in frame-wise action labelling, it is still necessary to provide supervision for \textit{every} video. 
\textbf{Unsupervised} approaches address action segmentation by combining clustering methods with temporal models (\eg, Hidden Markov Model)~\cite{sener2018unsupervised,kukleva2019unsupervised,li2021action}. While some work simply perform clustering on the input features which do not involve any learning and achieve very competitive results~\cite{sarfraz2021temporally,du2022fast}. Since no semantic labels are provided during learning, performances are evaluated based on the best Hungarian matching scores. One recent work ICC~\cite{singhania2021iterative} proposed a contrastive learning approach for unsupervised learning of frame-wise features. The learned representations are then adapted to a semi-supervised setting by learning a post-hoc linear classifier. The classifier incorporates the unlabelled data with na\"ive pseudo-labels, which weakens the overall contribution to the semi-supervised learning area.  

\subsection{Semi-supervised Learning}
Existing dominant approaches to image-based semi-supervised learning include consistency regularization~\cite{samuli2017temporal,tarvainen2017mean} and pseudo-labelling~\cite{lee2013pseudo}. Consistency regularization methods, such as Temporal Ensembling~\cite{samuli2017temporal} and Mean-teacher~\cite{tarvainen2017mean}, aim to learn the prediction consistency in different epochs or models with augmented inputs. Applying augmentations analogous to the image domain such as flipping, rotation, and transformation to videos for action segmentation is non-trivial as the inputs are pre-computed feature vectors. 

Pseudo-labelling methods generate labels for unlabelled data to guide learning~\cite{lee2013pseudo}. To generate pseudo-labels,~\cite{ding2019feature} leverages the sample similarity in the feature space to assign soft labels, whereas ~\cite{iscen2019label} implements a graph-based label propagation framework. Some researchers~\cite{cho2022pixel,wang2021self} have attempted to apply semi-supervised learning to video tasks by adapting image-based techniques to take video input. 
However, little has been done to evaluate the effectiveness of semi-supervised learning in temporal action segmentation. To address this research gap, we propose two novel loss functions designed based on the observation of two unique properties of procedural task videos.

\section{Method}
\subsection{Preliminaries}

We denote a labelled sample video sequence of temporal length $T$ as $\{(x^t, y^t)\}_{t=1}^{T}$, where $x^t$ is the video frame feature indexed at time $t$ and $y^t$ is its semantic action label. In a semi-supervised scenario, a labelled set $\mathcal{D}_L=\{(x_i,y_i)\}_{i=1}^N$ of $N$ labelled videos and an unlabelled set $\mathcal{D}_U=\{(x_j)\}_{j=1}^M$ of $M$ videos are given, where $M\gg N$. 
For every video in the small labelled set $\mathcal{D}_L$, each frame has a label from one of 
$K$ classes, \ie, $y_i^t \in \{1, 2,\dots,K\}$.  
The complete training set is denoted by $\mathcal{D}=\mathcal{D}_L+\mathcal{D}_U$. 

To learn a semi-supervised action segmentation model $\mathcal{M}$ parameterised by $\theta$, we use the labelled and unlabelled videos with the following objective:
\begin{equation}\label{eq:overall}
\min_\theta \sum_{(x,y)\in\mathcal{D}_L}\mathcal{L}^{L}(x,y;\theta) + \alpha \sum_{(x)\in \mathcal{D}_U} \mathcal{L}^{U} (x;\theta) + \beta \sum_{(x)\in \mathcal{D}}\mathcal{R}^D (x;\theta),
\end{equation}
where $\mathcal{L}^L$ denotes a supervised loss (Sec.~\ref{subsec:sup}), $\mathcal{L}^U$ denotes an unsupervised loss, and $\mathcal{R}^D$ is some regularization loss (Sec.~\ref{subsec:sup}), weighted by hyperparameters 
$\alpha,\beta \in \mathbb{R}_{>0}$. 
In the above objective, formulating the unsupervised loss $\mathcal{L}_{\text{u}}$ is vital for effective semi-supervised learning. In this work, we designed two novel loss functions for unlabelled data, \ie~action affinity loss (Sec.~\ref{subsec:aff}) and action continuity loss (Sec.~\ref{subsec:cont}), each attending to one characteristic we observed in procedural videos. 

\subsection{Supervised Temporal Action Segmentation}\label{subsec:sup}
Existing supervised approaches~\cite{farha2019ms,lea2017temporal,singhania2021coarse} take sequences of video frames as input and predict frame-wise action labels as a classification task. For the labelled data, we follow the same scheme to train the segmentation model $\mathcal{M}$ that estimates frame-wise action probabilities $p^t(k) = \mathcal{M}(x^t)$ with the classification loss formulated as:
\begin{equation}\label{eq:clsl}
    \mathcal{L}_{\text{cls}}^L = \frac{1}{T} \sum_t -\log(p^t(y^t))
\end{equation}
where $p^t\in \mathbb{R}^K$ is the estimated action class probability for frame $x^t$. 
It is also common to apply a smoothing loss with threshold $\tau$ to encourage smooth transitions between frames: 
\begin{equation}\label{eq:tmse}
    \mathcal{L}_{\text{sm}} = \frac{1}{TK}\sum_{t,k}\tilde{\Delta}_{t,k}^2, \;\;\;
    \tilde{\Delta}_{t,k} = \begin{cases}
    \Delta_{t,k} &: \Delta_{t,k} \le \tau\\
    \tau &: \text{otherwise}
    \end{cases},
\end{equation}
\begin{equation}
    \Delta_{t,k} = \left|\log p^t(k) - \log p^{t-1}(k)\right|.
\end{equation}
We follow~\cite{farha2019ms} and set $\tau$ to 4. 

\subsection{Action Affinity}\label{subsec:aff}
Videos performing the same (procedural) activity will share the same or a similar set of composing actions. We assume we can find similar videos that match and share resembling temporal proportions. 
With this motivation in mind, we define for a labelled video $i$ a video-level representation based on the action frequency: 
\begin{equation}
    q_i(k) = \frac{1}{T_i}\sum_{t}^{T_i} \mathbbm{1} (y_i^t == k); \quad k \in [1,\dots,K]
\end{equation}
For an unlabelled video $j$, we define a soft action frequency based on the network prediction outputs:
\begin{equation}\label{eq:unlabelled}
    p_j(k) = \frac{1}{T_j} \sum_t^{T_j} p_j^t(k);\quad k \in [1,\dots,K].
\end{equation}

\noindent Unless explicitly stated otherwise, we denote $q$, indexed by $i$, for labelled videos and $p$ with index $j$ for unlabelled videos.

\textbf{Anchor Association.} We want to provide action-level supervision for unlabelled videos by finding their most similar peers from the labelled set. Given some distance function $d(\cdot)$, we refer to a labelled video $i$ as an anchor $a_j$ for a unlabelled video $j$ if their action frequencies are the closest amongst the entire labelled set, \ie
\begin{equation}\label{eq:asso}
    a_j = q_{i^*}, \quad i^*=\argmin_i d(q_i, p_j)
\end{equation} 

\textbf{Affinity Loss.} Formally, we use the Kullback–Leibler~(KL) divergence as the distance criterion and define our action affinity loss as the affinity between the best matched pairs $(p_j, a_j)$, which is also the minimum distance between $p$ over the entire labelled set:
\begin{equation}\label{eq:kl}
    \mathcal{L}_{\text{aff}} = \sum_k a_j(k) \log\left(\frac{a_j(k)}{p_j(k)}\right) =  \min_{i} \sum_k q_i(k) \log\left(\frac{q_i(k)}{p_j(k)}\right)
\end{equation}
\begin{figure}[tb]
    \centering
    \begin{overpic}[width=0.5\linewidth]{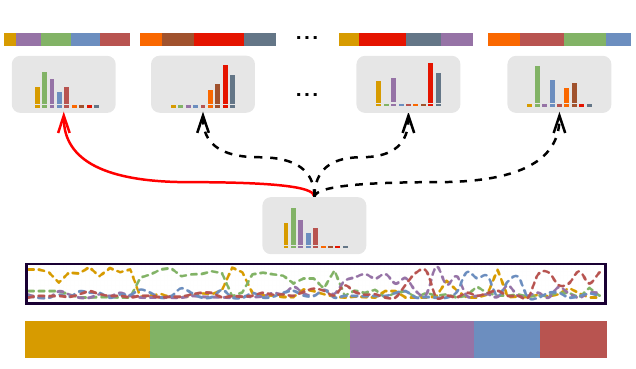}
    \put(-1,3){\rotatebox[origin=c]{90}{\tiny GT}}
    \put(-2,12){\tiny $p_j^t$:}
    \put(14,36.5){\tiny $q_1$}
    \put(35,36.5){\tiny $q_2$}
    \put(68,36.5){\tiny $q_3$}
    \put(92,36.5){\tiny $q_4$}
    \put(53,22){\tiny $p_j$}
    \put(4,27){\rotatebox[origin=c]{0}{\tiny anchor}}
    \put(4,24){\rotatebox[origin=c]{0}{\tiny association}}
    \put(38,55){\tiny Labelled videos}
    \end{overpic}
    \caption{Action affinity loss overview. Action frequencies $q, p$ are first built for both labelled and unlabelled videos. The action affinity loss associates (red arrow) for $p_j$ its nearest anchor $q_1$ in labelled set and then imposes the action prior from $q_1$ on $p_j$ to supervise the learning. Our affinity loss allows variations of action ordering (green and purple segments in GT and $q_1$). }
    \label{fig:aff}
\end{figure}
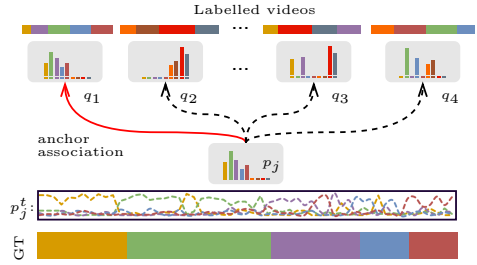

\noindent  Minimising the above action affinity loss imposes pair-wise action frequency prior from the labelled set; it guides network outputs to have similar action composition to labelled videos, which is especially important when using unlabelled sequences for training. Fig.~\ref{fig:aff} depicts our affinity loss. Empirically, this loss combined with a frame-wise entropy loss outperforms pseudo-labels (see. Sec.~\ref{subsec:ablation}).

\subsection{Action Continuity}\label{subsec:cont}

\begin{figure}[t]
    \centering
    \begin{overpic}[width=1\linewidth]{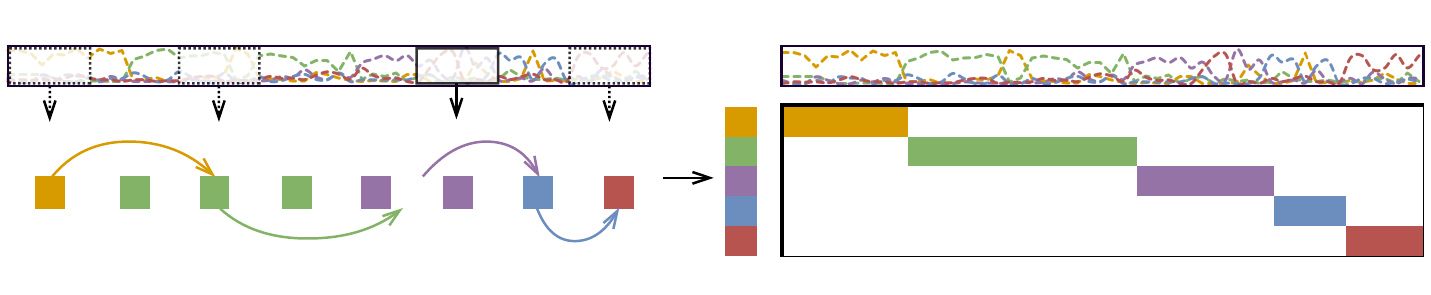}
    \put(10,1){\tiny Action Sequence Sub-sampling}
    \put(-2.5,15.5){\tiny $p_j^t$:} \put(51.5,15.5){\tiny $p_j^t$:}
    \put(2.5,15.5){$\omega$}\put(14.5,15.5){$\omega$}\put(31,15.5){$\omega$}\put(42,15.5){$\omega$}
    \put(87,11.5){\tiny Cost Matrix}
    \put(66,1){\tiny Dynamic Time Warping}
    \end{overpic}
    \caption{Action continuity loss overview. Given network predictions for an unlabeled video $p_j$, a sliding window sub-sampling is first performed to obtain an action sequence (order indicated by coloured arrows); the sequence is later compared against $p_j$ to construct a cost matrix. The action continuity loss is the average cost along the optimal assignment path (coloured segments in the cost matrix) found via dynamic time warping.}
    \label{fig:cont}
\end{figure}
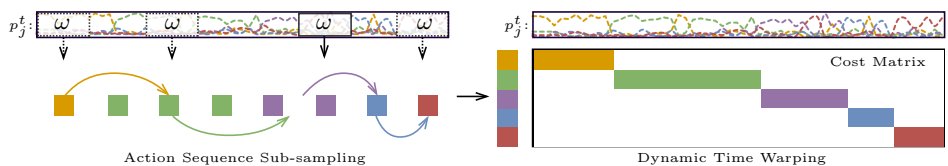

A simple way to generate pseudo-labels $\hat{y}$ for unlabelled video sequences is to use the class label with the maximum probability, which can be used to supervise the learning of unlabelled data with the classification loss:

\begin{equation}\label{eq:naive}
    \!\!\!\mathcal{L}_{\text{pse}} = -\frac{1}{T}\!\sum_t\! \log(p^t(\hat{y}^t)), \text{ where  } \hat{y}^t\!=\!\argmax_k p^t(k).
\end{equation}
However, such na\"ive pseudo-labels directly inferred from network outputs tend to be temporally over-fragmented~\cite{ishikawa2021alleviating}. This breaks the temporal continuity of actions, 
\ie, label changes should occur only at (true) action boundaries. 
To this end, we propose an action continuity loss to impose such action transition constraints. 
For an unlabelled video, this loss takes as input its frame-wise predictions, sub-samples in time and then estimates the learning objective via dynamic time warping as illustrated in Fig.~\ref{fig:cont}.

\textbf{Action Sequence Sub-sampling.} We first generate action candidates which have maximum average class probability within a sliding window of stride $\omega$, %
\begin{equation}
    o = \argmax_k \frac{1}{\omega}\sum_{t=t'}^{t'+\omega} p^t(k),\label{eq:o}
\end{equation}
where $t'$ is the previous temporal window location. Subsequently, we yield an ordered sequence with $\lceil\frac{T}{\omega}\rceil$ elements 
denoted as $\mathcal{O} = \{o^l\}_{l=1}^{\lceil\frac{T}{\omega}\rceil}$. This sequence can be further reduced in length by removing the adjacent action repetitions, \ie, $o^l = o^{l+1}$, to a length of $L$.

\textbf{Dynamic Time Warping.} 
Given an unlabelled video with its frame-wise action probabilities $p_j$ of length $T$ and its inferred action sequence $\mathcal{O}$ of length $L$ as described above, a cost matrix of alignment $\Delta = \{d(l,t)\}\in \mathbb{R}^{L\times T}$ can be constructed with some distance function $d(\cdot,\cdot)$. Using dynamic time warping, we find the best possible alignment $Y^*$ defined by the following objective:
\begin{equation}
   Y^* = \argmin_{Y} \langle Y, \Delta\rangle \label{eq:dtw}
\end{equation}
where $\langle\cdot,\cdot\rangle$ is the inner product and $Y \subset \{0,1\}^{L\times T}$ is a binary assignment matrix. $Y_{tl} = 1$ if frame $t$ has the label $o^l$ and $Y_{tl}=0$ otherwise. Eq.~\eqref{eq:dtw} is solved efficiently with dynamic programming. The label assignment $\Tilde{y}^t$ for $p^t$ can be then inferred by parsing $Y^*$:
\begin{equation}\label{eq:ty}
    \Tilde{y}^t = \sum_l o^l \mathbbm{1}(Y^*_{lt} == 1)
\end{equation}

\textbf{Continuity Loss.}
An intuitive way of forming the continuity loss is to take the optimal objective of dynamic time warping and minimise it:
\begin{equation}
    \mathcal{L}_{\text{cont}} = \langle Y^*,\Delta \rangle
\end{equation}
We achieve this by choosing a specific distance function $d$. With a slight abuse of notation, we denote the categorical label $o$ (Eq.~\eqref{eq:o}) as its one-hot embedding when written as $o(k)$ and designate the distance function as the KL divergence:
\begin{equation}\label{eq:dist}
    d(l,t) = KL(o^l||p^t) = \sum_k o^l(k) \log\left(\frac{o^l(k)}{p^t(k)}\right).
\end{equation}
If we replace the $o^l$ with the final assignment $\Tilde{y}^t$ for $p^t$ in Eq.~\eqref{eq:dist}, the cost for $p^t$ in the optimal $Y^*$ would become the negative log-likelihood $-\log(p^t(\Tilde{y}^t))$. 
Averaging the cost over the entire video sequence leads to our final action continuity loss formulation:
\begin{equation}\label{eq:clsu}
    \mathcal{L}_{\text{cont}} = \frac{1}{T} \min_Y \langle Y,\Delta \rangle = \frac{1}{T} \sum_t -\log(p^t(\Tilde{y}^t)).
\end{equation}
We note that with the KL divergence distance function, this continuity loss is consistent with the frame-wise classification loss enforcing the network predictions to approximate $\tilde{y}$, which is temporally continuous.

Adopting Eq.~\eqref{eq:clsl} for labelled data ($\mathcal{L}^L$), Eq.~\eqref{eq:kl} and Eq.~\eqref{eq:clsu} for the unlabelled data ($\mathcal{L}^U$), and Eq.~\eqref{eq:tmse} for the regularization ($\mathcal{R}^D$), we can rewrite Eq.~\eqref{eq:overall} as our semi-supervised learning objective with the following form:
\begin{equation}\label{eq:all}
    \mathcal{L} = \mathcal{L}_{\text{cls}}^L+\alpha \mathcal{L}_{\text{aff}}^{U} + \beta \mathcal{L}_{\text{cont}}^{U} + \gamma \mathcal{L}_{\text{sm}}^D
\end{equation}
where $\alpha,\beta,\gamma$ are trade-off parameters balancing the terms. The smoothing loss $L_{\text{sm}}^D$ is imposed on the full set of data.

\begin{figure}[tb]
    \subfigure[Standard One-hot Labels]{\label{fig:one-hot}
        \includegraphics{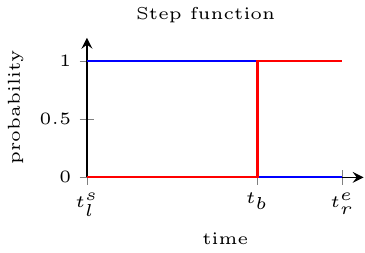}
    }
    \subfigure[Fixed-Duration Linear~\cite{ding2018weakly}]{\label{fig:fixed}
        \includegraphics{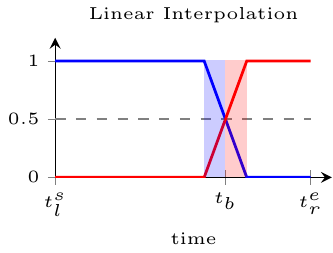}
    }
    \subfigure[ABS (Ours)]{\label{fig:abs}
    \includegraphics{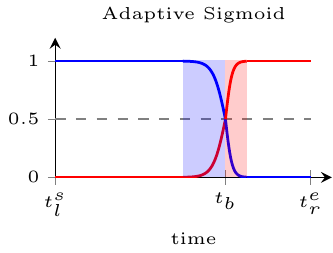}
    }
    \caption{Probability assignment approaches around the action boundary as a function of time. 
    Let $t_b$ denote the estimated boundary between the left action in $[t_l^s,t_b)$ and the right action $[t_b,t_r^e)$. The blue and red shaded segments denote the boundary vicinities $V_l$ and $V_r$. (a) The standard one-hot labels adopt a step function and assign hard action labels for all the frames. (b) The fixed-duration linear approach~\cite{ding2018weakly} mixes the action probabilities linearly with a fixed slope around the boundary. (c) ABS (Ours) uses a sigmoid function with a decay proportional to the action duration.
    }
    \label{fig:boundary}
\end{figure}

\subsection{Adaptive Boundary Smoothing (ABS)}
In our semi-supervised setting, the action boundaries of an unlabelled video inferred from the best possible assignment $Y^*$ or $\Tilde{y}$ may still be inaccurate. As such, we propose an adaptive boundary smoothing (ABS) technique to provide softer %
action boundary supervision for more robust and reliable learning. Boundary smoothing was initially proposed in~\cite{ding2018weakly} and has been explored in the weakly-supervised setting to improve the segmentation performance. 
Unlike~\cite{ding2018weakly}, which uses a fixed linear-interpolation scheme for smoothing, we use an adaptive scheme based on the estimated action duration. This allows us to elastically mix action probabilities for frames within the vicinity of the boundary.

\textbf{Duration Aware Boundary Vicinity.} Given left and right action segments ($\mathcal{S}_l: [t_l^s, t_l^e, y_l], \mathcal{S}_r: [t_r^s, t_r^e, y_r]$) consecutive in time, let $t^s, t^e$ denote the starting and ending timestamps of the action and $y \in [1,\dots, K]$ the corresponding semantic label.  The action boundary in between can be denoted as $t_b\!=\!t_r^s\!= \!t_l^e+1$. With a vicinity parameter $v \in [0,0.5]$, we define the boundary vincinities or ranges $V_l$ and $V_r$ for the left and right actions respectively: 
\begin{equation}\label{eq:v}
    V_l = [t_b - (t_b-t_l^s) * v,  t_b), \quad \text{and} \quad  V_r=[t_b, t_b+(t_r^e-t_b)*v).
\end{equation}

\textbf{Adaptive Sigmoid.} Within each action boundary vicinity $V$, we utilize an adaptive sigmoid function to assign mixed probabilities. For a frame within the left boundary vicinity, \ie, $ t\in V_l$, its smoothed probabilities for two action classes ($y_l,y_r$) are written as:
\begin{equation}\label{eq:prob}
    y^t(y_l) = \frac{1}{1+e^{-\frac{\epsilon}{|V_l|}(t-t_b)}},\quad \text{and} \quad  y^t(y_r) = 1-y^t(y_l)
\end{equation}
where $\epsilon$ is a predefined parameter which is set to 5 to ensure that the furthest frame to the boundary in a vicinity set has close to 1 probability for the action label of the segment it belongs to. $|V|$ denotes the temporal length of $V$. Probability assignment for $V_r$ is identical to Eq.~\eqref{eq:prob} but with $y_r$ and $y_l$ changed.

ABS can be efficiently incorporated in our approach by replacing the one-hot action probability within each boundary vicinity from $\Tilde{y}$ (Eq.~\eqref{eq:ty}) with the above mixed probabilities. With $v=0$, our ABS degenerates into the one-hot setting. Fig.~\ref{fig:boundary} compares three types of action probability assignments around the action boundaries. One-hot labels (Fig.~\ref{fig:one-hot}) are standard in practice and assume rigid action boundaries. Fixed-duration linear~\cite{ding2018weakly} (Fig.~\ref{fig:fixed}) softens the boundary with linearly interpolated action probabilities in a fix-sized temporal window.
In contrast, our proposed ABS approach (Fig.~\ref{fig:abs}) allows the corresponding action probabilities of vicinity frames from a longer action segment to have a faster-descending speed when approaching the boundary and vice-versa. Smoothing in a larger vicinity of longer segments provides more training samples for shorter segments, while smoothing in a smaller vicinity of shorter segments helps preserve more high confident middle frames for learning.

\section{Experiments}

\subsection{Datasets, Protocols and Evaluation}

\textbf{Datasets.} We conducted our experiments on the three benchmark datasets.
\textbf{Breakfast Actions}~\cite{kuehne2014language} comprises in total 1712 videos performing ten different activities with 48 actions. On average, each video contains six action instances. \textbf{50Salads}~\cite{stein2013combining} has 50 videos with 17 action classes. \textbf{GTEA}~\cite{fathi2011learning} contains 28 videos of seven kitchen activities composing 11 different actions. \\

\noindent \textbf{Protocols.} We used the standard train-test splits for each dataset; we randomly selected 5\%, 10\% of the training set as the labelled set $\mathcal{D}_L$ and regarded the remaining training videos as the unlabelled set $\mathcal{D}_U$. The labelled set was ensured to contain at least one segment instance of each action. For 50Salads and GTEA, 3 and 5 videos were sampled in place of 5\% and 10\% of labelled data as the datasets are relatively small.  \\

\noindent \textbf{Evaluation.} 
We adopted the same evaluation metrics as fully-supervised action segmentation and reported frame-wise accuracy (Acc), segmental edit score (Edit), and segmental F1 score with varying overlap thresholds 10\%, 25\%, and 50\%. For all datasets, we randomly sampled five labelled subsets from the original training data. We cross-validated over the standard splits and reported the average over the splits across the five runs. 

\subsection{Implementation Details}
We use the multi-stage temporal convolutional network (MS-TCN)~\cite{farha2019ms} as the backbone segmentation model $\mathcal{M}$. Our model was first warmed up with only labelled data for 30 epochs, and then unlabelled data was incorporated for another 20 epochs. The initial learning rate was set as $5e^{-4}$. We used the Adam optimiser, with weights settings of $\alpha=0.1$, $\beta=0.01$, and $\gamma=0.15$, as per~\cite{farha2019ms}.  
The action sequence sub-sampling stride $\omega$ was set to 20. We set the vicinity parameter $v=0.05$ for all three datasets. 

\subsection{Effectiveness}
Table~\ref{tab:main} reports the improvements of our method compared with the supervised baseline (\textit{Base}) and na\"ive pseudo-labelling approach (\textit{Pseudo}) on three benchmarks. \textit{Base} is trained with only labelled data while \textit{Pseudo} assigns pseudo-labels and trains with $\mathcal{L_\text{pse}}$ (Eq.~\eqref{eq:naive}). In both the 5\% and 10\% settings, our model consistently outperformed the \textit{Base} model by a large margin. Specifically, on the 50Salads dataset, the accuracy of our model increased by 26\% (from 32.0\% $\rightarrow$ 58.0\%). The overall increase in performance across datasets was greater when more labelled data (5\%$\rightarrow$10\%) was provided. It is noteworthy that on 50Salads with 5\% labelled data, the segmentation performances for \textit{Pseudo} are lower than the \textit{Base} by around 3\%, which shows that the model overfitted to inaccurate pseudo-labels, likely due to confirmation bias. On the contrary, our proposed approach can still significantly boost the accuracy performance by a large gain of 24.2\%. This verifies the effectiveness of the valuable action affinity prior information inferred from the rarely few labelled video samples.  

\begin{table}[tb]
    \centering
    \caption{Performance of our proposed approach on three benchmark datasets}\label{tab:main}
    \scalebox{0.88}{
        \begin{tabular}{c|l|ccccc|ccccc|ccccc}
        \hline
        \multirow{2}{*}{\%$D_L$} & \multirow{2}{*}{Method} & \multicolumn{5}{c|}{Breakfast} & \multicolumn{5}{c|}{50Salads} & \multicolumn{5}{c}{GTEA} \\ \cline{3-17}
         &  & \multicolumn{3}{c}{F1@\{10, 25, 50\}} & Edit & Acc & \multicolumn{3}{c}{F1@\{10, 25, 50\}} & Edit & Acc & \multicolumn{3}{c}{F1@\{10, 25, 50\}} & Edit & Acc \\ \cline{1-17}
        \multirow{4}{*}{5} & Base & 36.7 & 28.4 &	19.5 & 37.5 & 28.2 & 26.8 & 19.7 & 11.5 & 26.1 & 28.1 & 29.9 & 25.8 & 14.8 & 31.0 & 37.2\\
         & Pseudo & 40.2 & 28.5 & 20.1 & 41.3 & 20.9 & 22.6 & 17.0 & 12.1 & 22.0 & 24.0 &  48.4 & 42.3 & 30.2 & 45.4 & 48.1 \\ 
         & Ours & 44.5 & 35.3 & 26.5 & 45.9 & 38.1 & 37.4 & 32.3 & 25.5 & 32.9 & 52.3 & 59.8 & 53.6 & 39.0 & 55.7 & 55.8 \\ 
        \rowcolor{gray!30}
         & Gain & \textbf{7.8} & \textbf{6.9} & \textbf{7.0} & \textbf{8.4} & \textbf{9.9} & \textbf{10.6} & \textbf{12.6} & \textbf{14.0} & \textbf{6.8} & \textbf{24.2} & \textbf{29.9} & \textbf{27.8} & \textbf{24.2} & \textbf{24.7} & \textbf{18.6} \\
         \hline
        \multirow{4}{*}{10} & Base & 46.8 & 41.1 & 29.2 & 50.9 & 37.1 & 27.6 & 24.3 & 16.0 & 27.4 & 32.0 & 38.1 & 29.6 & 15.3 & 39.6 & 41.1\\
         & Pseudo & 49.3 & 44.8 & 33.9 & 49.7 & 40.2 & 36.2 & 32.4 & 24.5 & 33.5 & 41.1 & 65.5  & 60.7 & 45.8 & 59.9 & 57.9 \\
         & Ours & 56.9 & 51.3 & 39.0 & 57.7 & 49.5 & 47.3 & 42.7 & 31.8 & 43.6 & 58.0 & 71.5 & 66.0 & 52.9 & 67.2 & 62.6 \\
         \rowcolor{gray!30}
         & Gain &\textbf{10.1} & \textbf{10.2} & \textbf{9.8} & \textbf{6.8} & \textbf{12.4} & \textbf{19.7} & \textbf{18.4} & \textbf{15.8} & \textbf{16.2} & \textbf{26.0} & \textbf{33.4} & \textbf{36.4} & \textbf{37.6} & \textbf{27.6} & \textbf{21.5} \\\hline
        \end{tabular}
    }
\end{table}

\begin{table}[t]
    \begin{minipage}{.5\textwidth}
        \centering
        \caption{Comparison of frame accuracy boost between different dataset variances}\label{tab:covarinace}
        \begin{tabular}{l|c|ccc}
        \hline
        & labelled &50Salads  & GTEA & Breakfast\\ \hline
        $var$ & - & \cellcolor{gray!30}8$e^{-4}$  & \cellcolor{gray!20}3$e^{-3}$ & \cellcolor{gray!10}6$e^{-3}$ \\ 
        Gain & 5\% & \cellcolor{gray!30}24.2 & \cellcolor{gray!20}18.6 & \cellcolor{gray!10}9.9 \\
        Gain & 10\% & \cellcolor{gray!30}26.0 & \cellcolor{gray!20}21.5 &  \cellcolor{gray!10}12.4\\
        \hline
        \end{tabular}
    \end{minipage}
    \hfill
    \begin{minipage}{0.5\textwidth}
        \centering
        \caption{Effect of activity labels on Breakfast (5\%)}\label{tab:ca}
        \begin{tabular}{l|ccccc}
        \hline
         & \multicolumn{3}{c}{F1@\{10,25,50\}} & Edit & Acc \\\hline
        w/o activity & 44.5& 35.3 & 26.5 & 45.9 & 38.1 \\
        w/ activity & 56.6 & 49.3 & 35.8 & 59.4 & 56.6\\
        \rowcolor{gray!30}
        Gain & \textbf{12.1} & \textbf{14.0} & \textbf{9.3}  & \textbf{13.5} & \textbf{18.5}\\\hline
        \end{tabular}
    \end{minipage}
\end{table}

\textbf{Affinity Association.}
Amongst all three datasets in Table~\ref{tab:main}, the increase in Acc performance was the greatest on 50Salads and the least on Breakfast. We speculate that this is related to how accurate the affinity association (Eq.~\eqref{eq:asso}) is in finding the anchor videos from the labelled set. 
We validated this by calculating the total variance amongst the full set $\mathcal{D}$. The total variance is defined as the trace of the action frequency covariance matrix $\mathcal{C}\in\mathbb{R}^{K\times K}$ normalized by the number of actions, \ie, $var = \text{tr}(\mathcal{C}) / K$. The lower the variance, the more likely our affinity loss provided accurate supervision. The extreme case where the full set of videos share the identical action composition and frequency ($var\!=\!0$) guarantees that the supervision by affinity loss is always accurate and precise. Table~\ref{tab:covarinace} verifies that datasets with smaller variances had higher accuracy gains. Breakfast has the largest variance because it has 10 activities with slightly overlapping composing actions.  By extension, its dispersed action frequency representations would cause the variance for a single action to be high.

We also evaluated our approach on the Breakfast dataset {with video-level activity labels provided for all videos} and reported the results in Table.~\ref{tab:ca}. In this setting, we only searched for anchors from labelled videos with the same activity label. As we can see, when activity labels were given, the performance had a striking improvement of 18.5\% in accuracy. This is because these high-level labels excluded the incorrect anchor associations across two different activities. Such improvement validates our affinity observation in the same activity videos.

\subsection{Ablation Studies}\label{subsec:ablation}
\begin{figure*}
\begin{minipage}{0.45\textwidth}
        \centering
        \captionof{table}{Loss function ablation study on Breakfast (10\%)}\label{tab:varients}
        \scalebox{0.8}{
        \begin{tabular}{cccccccccc}
        \hline
        $\mathcal{L}_{\text{cls}}$ &$\mathcal{L}_{\text{pse}}$ & $\mathcal{L}_{\text{aff}}$ & $\mathcal{L}_{\text{cont}}$   & ABS & \multicolumn{3}{l}{F1@\{10, 25, 50\}} & Edit & Acc  \\ \hline
        \checkmark &&&      & & 47.9       & 40.6       & 28.6      & 51.8 & 36.8 \\
        \checkmark &\checkmark&       &&   & 49.3       & 44.8      & 33.9     & 49.7 & 40.2\\ 
        \checkmark&& \checkmark&    & & 52.0       & 46.5       & 34.3      & 53.4 & 44.0 \\
        \checkmark&\checkmark&\checkmark& && 54.1       & 46.7       & 34.9      & 54.1 & 47.8 \\
        \checkmark& & \checkmark &\checkmark& & 53.8 & 50.1 & 37.6 & 56.6 & 49.2\\
        \rowcolor{gray!30}
        \checkmark& & \checkmark &\checkmark& \checkmark & \textbf{56.9} & \textbf{51.3} & \textbf{39.0} & \textbf{57.7} &\textbf{49.5}\\\hline
        \end{tabular}}
    \end{minipage}
\hfill
\begin{minipage}{0.45\textwidth}
    \centering
    \includegraphics{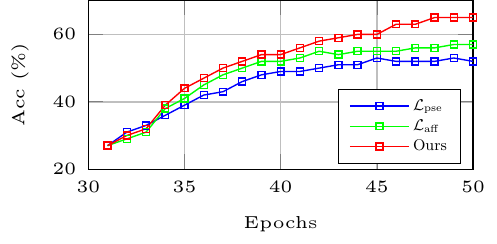}
    \captionof{figure}{Pseudo-label accuracy against training epochs on GTEA (10\%).}\label{fig:pseudo}
\end{minipage}
\end{figure*}

\textbf{Loss Functions and ABS.}
Table~\ref{tab:varients} reports the ablation study results on different variants of loss functions and ABS. The first row is the baseline model trained with only labelled data and $\mathcal{L}_{\text{cls}}$. 
Results for na\"ive pseudo-labelling loss $\mathcal{L}_{\text{pse}}$ (Eq.~\eqref{eq:naive}) in the second row show 
a mild increase in F1 scores and accuracy compared to the baseline. While more unlabelled data was accessible for learning, using them in the form of pseudo-labels brought little advantage. On the other hand, our proposed action affinity loss $\mathcal{L}_{\text{aff}}$ (third row) surpassed the pseudo-labelling counterpart by a margin of around 2\% on all metrics. We imposed an extra frame-wise entropy loss formulated as $-\sum_k p_i^t(k) \log p_i^t(k)$ in this variant, which forced the network to produce confident frame-wise predictions as the affinity loss does not provide frame-level supervision.
The combination of action affinity and na\"ive pseudo-labelling (fourth row) further enhanced the performance. Such improvements show that our affinity loss $\mathcal{L}_{\text{aff}}$ can improve the quality of pseudo-labels, which we will evaluate in the following text. Our model combining $\mathcal{L}_{\text{aff}}$ and $\mathcal{L}_{\text{cont}}$ achieved better performance than all the above variants. Lastly, as indicated by the last row, the integration of our proposed ABS further boosted the segmentation performance on Breakfast.  

\textbf{Pseudo-labels.} We studied the quality of estimated action classes on the unlabelled data $\mathcal{D}_U$. %
Fig.~\ref{fig:pseudo} shows a plot of the pseudo-label accuracy between training epochs. In the first epoch that unlabelled data was incorporated for training, all variants achieved the same accuracy, but the scores diverged as the training progressed. Imposing $\mathcal{L}_{\text{aff}}$ led to better accuracy compared to $\mathcal{L}_{\text{pse}}$, while our full loss formulation predicted the most accurate pseudo-labels for unlabelled data.

\begin{table}[tb]
\begin{minipage}{0.5\textwidth}
\centering
\caption{Sub-sampling stride $\omega$ on Breakfast (10\%)}\label{tab:omega}
\begin{tabular}{@{}l|cccccc@{}}
\hline
$\omega$    & 10   & 15   & 20   & 25   & 30 & 60  \\ \hline
Acc & 48.5 & 48.8 &\cellcolor{gray!30}\textbf{49.5} & 49.0 & 47.9 & 45.6\\ \hline
\end{tabular}
\end{minipage}
\hfill
\begin{minipage}{.5\textwidth}
\centering 
\caption{Effect of $\alpha, \beta$ on GTEA (10\%)}\label{tab:hyper}
\begin{tabular}{lcccc}
\hline
&&$\alpha$&&\\\hline
$\beta$    & 1    & 0.1  & 0.01 & 0.001 \\ \hline
1     & \cellcolor{gray!5}57.0 & \cellcolor{gray!16}59.3 & \cellcolor{gray!13}58.7 & \cellcolor{gray!1}54.5  \\ 
0.1   & \cellcolor{gray!15}58.9 & \cellcolor{gray!20}60.3 &  \cellcolor{gray!25}61.5 & \cellcolor{gray!10}57.9  \\ 
0.01  & \cellcolor{gray!15}59.2 & \cellcolor{gray!40}\textbf{62.6} & \cellcolor{gray!30}61.9 & \cellcolor{gray!10}58.0  \\ 
0.001 & \cellcolor{gray!10}58.3 &  \cellcolor{gray!30}61.4 &  \cellcolor{gray!20}60.5 & \cellcolor{gray!6}57.3  \\ \hline
\end{tabular}
\end{minipage}
\end{table}

\textbf{Sub-sampling Stride $\omega$.} 
Table~\ref{tab:omega} shows the accuracy changes with respect to the sub-sampling stride $\omega$ in Eq.~\eqref{eq:o} on Breakfast with 10\% labelled data. The frame accuracy fluctuates around 49\% with small strides, but when the stride becomes too large, \eg, $\omega\!=\!60$, the performance dropped by a margin of 3.9\% as some actions are likely to be fully skipped during sub-sampling. The best accuracy of 49.5\% was achieved when $\omega\!=\!20$. 

\textbf{Loss Hyperparameters.} The effect of hyperparameters is presented in Table~\ref{tab:hyper}. A very small weight on our affinity loss ($\alpha=0.001$) led to the lowest performance, as indicated by the last column. Increasing $\alpha$ boosted the performance, which shows that the action priors from affinity loss is vital. The comparison between the rows indicates that a large weight for the action continuity loss, \eg, $\beta=1$, caused the model to overfit to the inaccurate pseudo-labels and produced inferior results since it also provided frame-wise pseudo-supervision. 
The overall best performance arrived at 62.6\% with $\alpha=0.1, \beta=0.01$.

\textbf{Vicinity Parameter $v$.} Table~\ref{tab:v} compares ABS against One-hot and Fixed-duration linear~\cite{ding2018weakly}. ASB with $v=0.05$ enhanced the segmentation results by 1-2\% compared to the baseline One-hot ($v=0$). Fixed-duration linear~\cite{ding2018weakly} was also helpful, but the performance gain was only marginal. Setting $v=0.1$ doubles the vicinity, which experienced a performance drop compared to $v=0.05$; this likely indicates that the smoothing range is too large.

\textbf{ABS for Supervised Learning.}  Given that ABS is a general smoothing technique, we further integrated ABS with the fully-supervised setting and report the results in Table~\ref{tab:boundary_full}. A consistent increase in segmentation performance compared to the baseline was observed across all three datasets. Also, the relatively large improvements were made in segmental metrics (F1 and Edit scores).

\begin{table}[tb]
\centering
\caption{Effectiveness of ABS for fully-supervised action segmentation}\label{tab:boundary_full}
\scalebox{1}{
\begin{tabular}{c|ccccc|ccccc|ccccc}
\hline
         & \multicolumn{5}{c|}{Breakfast}& \multicolumn{5}{c|}{50Salads}& \multicolumn{5}{c}{GTEA} \\\hline
         & \multicolumn{3}{c}{F1@\{10,25,50\}} & Edit & Acc  & \multicolumn{3}{c}{F1@\{10,25,50\}} & Edit & Acc  & \multicolumn{3}{c}{F1@\{10,25,50\}} & Edit & Acc  \\\hline
    Base     & 63.2       & 57.7       & 45.6      & 65.5 & 65.1 & 66.8       & 63.7       & 55.2      & 59.8 & 78.2 & 84.9       & 82.4       & 67.6      & 79.7 & 76.6 \\
    +ABS    & 71.3       & 65.9       & 52.2      & 71.8 & 68.9 & 72.5       & 70.1       & 61.8      & 66.8 & 79.8 & 87.6       & 85.4       & 71.7      & 82.8 & 77.4 \\
    \rowcolor{gray!30}
    Gain &  \textbf{8.1}  &  \textbf{8.2}  &\textbf{ 6.6} & \textbf{6.3} & \textbf{3.8} & \textbf{5.7} &  \textbf{6.4} & \textbf{6.6} & \textbf{7.0} & \textbf{1.6}  &  \textbf{2.7 }& \textbf{3.0} & \textbf{4.1}  &\textbf{3.1}&\textbf{0.8}  \\\hline
    \end{tabular}}

\end{table}

\begin{table}[tb]
    \begin{minipage}{.47\textwidth}
        \caption{Comparison of vicinity $v$ \\on Breakfast (10\%)}\label{tab:v}
        \scalebox{0.85}{
        \begin{tabular}{l|ccccc}
        \hline
        Method & \multicolumn{3}{l}{F1@\{10,25,50\}} & Edit & Acc \\\hline
        One-hot($v=0$) & 53.8 & 50.1 & 37.6 & 56.6 & 49.2 \\
        Fixed-duration~\cite{ding2018weakly} & 54.7 & 50.5 & 38.1 & 56.9 & 49.1 \\
        \rowcolor{gray!30}
        $v=0.05$ &\textbf{56.9} & \textbf{51.3} & \textbf{39.0} & \textbf{57.7} & \textbf{49.5} \\
        $v=0.1$ & 55.1 & 50.9 & 37.9 & 57.0 & 48.9 \\\hline
        \end{tabular}}
    \end{minipage}
    \begin{minipage}{.5\textwidth}
    \centering
        \caption{Accuracy performance comparison with approaches under various supervisions, * denotes test data used for training}\label{tab:sotaa}
        \scalebox{.9}{
        \begin{tabular}{c|l|ccc}
        \hline
         & Method & Breakfast & 50salads & GTEA \\ \hline
        \multirow{3}{*}{\rotatebox[origin=c]{90}{Full}} & MSTCN~\cite{farha2019ms} & 65.1 & 78.2 & 76.6 \\
         & SSTDA~\cite{chen2020action}* & \textbf{70.2} & \textbf{83.2} & 79.8 \\
         & Ours (100\%)* & 69.3 & 82.5 & \textbf{80.4} \\ \hline
        \multirow{2}{*}{\rotatebox[origin=c]{90}{Weak}} & Timestamp~
        \cite{li2021temporal}& 64.1 & 75.6 & 66.4 \\
         & SSTDA~\cite{chen2020action} (65\%)* & 65.8 & 80.7 & 75.7 \\ \hline
        \multirow{3}{*}{\rotatebox[origin=c]{90}{Semi}} & Ours (5\%) & 38.1 & 52.3 & 55.8 \\
         & Ours (10\%) & 49.5 & 58.0 & 62.6 \\
         
         & \cellcolor{gray!30}Ours (50\%) &\cellcolor{gray!30} \textbf{63.9} &\cellcolor{gray!30} \textbf{78.8} & \cellcolor{gray!30}\textbf{77.9} \\ \hline
        \end{tabular}}
    \end{minipage}
\end{table}
\subsection{Comparison to State-of-the-Art Approaches}
We list in Table~\ref{tab:sotaa} relevant state-of-the-art approaches adopting MS-TCN~\cite{farha2019ms} or its variants as the backbone for a fair comparison. We did not include ICC~\cite{singhania2021iterative} as their approach cannot work with the MS-TCN architecture. For the ``Full'' comparison, we followed SSTDA~\cite{chen2020action} and applied our semi-supervised method to 100\% labelled data. We used the test data as the unlabelled set and achieved comparable performance. The frame accuracy of Timestamp~\cite{li2021temporal}, which uses per-segment supervision for all video samples, is close to fully-supervised MS-TCN~\cite{farha2019ms} except on GTEA. With 50\% labelled data, our approach managed to achieve comparable or better performance compared to Timestamp~\cite{li2021temporal}, MS-TCN~\cite{farha2019ms} as well as SSTDA~\cite{chen2020action} using a larger percentage (65\%) of labelled data. 

\section{Conclusion}
Procedural videos performing the same tasks exhibit affinity in action composition and continuity in action duration. Based on these unique characteristics, we proposed two novel loss functions for the semi-supervised temporal action segmentation task. The action affinity loss harnessed the action priors from the labelled set to supervise the unlabelled data. The action continuity loss function sub-sampled action sequence to enforce the temporal continuity of actions and provided frame-wise supervision. Furthermore, we proposed an adaptive boundary smoothing technique for more robust action boundaries.
Our approach significantly improves the segmentation performance with a very small amount (5\% and 10\%) of labelled data and reaches comparable performance to the full supervision methods with 50\% labelled videos. 

\noindent \textbf{Acknowledgements:} This research is supported by the National Research Foundation, Singapore under its NRF Fellowship for AI (NRF-NRFFAI1-2019-0001). Any opinions, findings and conclusions or recommendations expressed in this material are those of the author(s) and do not reflect the views of National Research Foundation, Singapore.

\bibliographystyle{splncs04}
\bibliography{main.bib}
\end{document}